\documentclass[lettersize,journal]{IEEEtran}
\usepackage{amsmath,amsfonts}
\usepackage{algorithmic}
\usepackage{array}
\usepackage[caption=false,font=normalsize,labelfont=sf,textfont=sf]{subfig}
\usepackage{textcomp}
\usepackage{stfloats}
\usepackage{url}
\usepackage{verbatim}
\usepackage{multirow}
\usepackage{graphicx}
\def\BibTeX{{\rm B\kern-.05em{\sc i\kern-.025em b}\kern-.08em
    T\kern-.1667em\lower.7ex\hbox{E}\kern-.125emX}}
\usepackage{balance}

\begin{document}

\title{Exploiting Object-based and Segmentation-based Semantic Features for Deep Learning-based Indoor Scene Classification}
%\title{GOS2F2App: Exploiting Semantic-based Features for Deep Learning-based Indoor Scene Classification}

\author{Ricardo~Pereira$^1$,
        Lu\'{i}s~Garrote$^1$,
        Tiago~Barros$^1$,
        Ana~Lopes$^{1,2}$,
        and Urbano~J.~Nunes$^1$
        %and~Jane~Doe,~\IEEEmembership{Life~Fellow,~IEEE}
        
%\thanks{All authors are with the University of Coimbra, Institute of Systems and Robotics, Department of Electrical and Computer Engineering XXXX}}
\thanks{$^{1}$Authors are with the University of Coimbra, Institute of Systems and Robotics, Department of Electrical and Computer Engineering, Portugal. Emails: \footnotesize\{ricardo.pereira,~garrote,~tiagobarros,~anacris,~urbano\}@isr.uc.pt.}
\thanks{$^{2}$Author is also with the Polytechnic Institute of Tomar, Portugal.}
}

%M. Shell was with the Department of Electrical and Computer Engineering, Georgia Institute of Technology, Atlanta, GA, 30332 USA e-mail: (see http://www.michaelshell.org/contact.html).}% <-this % stops a space \thanks{J. Doe and J. Doe are with Anonymous University.}% <-this % stops a space
%\thanks{Manuscript received April 19, 2005; revised August 26, 2015.}}

%\author{IEEE Publication Technology Department
%\thanks{Manuscript created October, 2020; This work was developed by the IEEE Publication Technology Department. This work is distributed under the \LaTeX \ Project Public License (LPPL) ( http://www.latex-project.org/ ) version 1.3. A copy of the LPPL, version 1.3, is included in the base \LaTeX \ documentation of all distributions of \LaTeX \ released 2003/12/01 or later. The opinions expressed here are entirely that of the author. No warranty is expressed or implied. User assumes all risk.}}

%\markboth{Journal of \LaTeX\ Class Files,~Vol.~18, No.~9, September~2020}%
%{How to Use the IEEEtran \LaTeX \ Templates}

\maketitle

\begin{abstract}
%Indoor scenes are usually characterized by scattered objects and their relationships, which turns the indoor scene classification task into a challenging computer vision task. 
%While object detection techniques provide the 2D location of objects allowing to obtain spatial distributions between objects, semantic segmentation techniques provide pixel-level information that allows to obtain, at a pixel-level, a spatial distribution and shape-related features of the segmentation-categories. 
%Hence, a novel approach that uses a semantic segmentation mask to provide Hu-moments-based segmentation-categories' shape characterization, designated by Segmentation-based Hu-Moments Features (SHMFs), is proposed. Moreover, a three-main-branch network, designated by GOS$^2$F$^2$App, that exploits deep-learning-based global features, object-based features, and semantic segmentation-based features is also proposed. GOS$^2$F$^2$App was evaluated in two indoor scene benchmark datasets: SUN RGB-D and NYU Depth V2, where, to the best of our knowledge, state-of-the-art results were achieved on both datasets, presenting evidences of the effectiveness of the proposed approach.

Indoor scenes are usually characterized by scattered objects and their relationships, which turns the indoor scene classification task into a challenging computer vision task. Despite the significant performance boost in classification tasks  achieved in recent years, provided by the use of deep-learning-based methods, limitations such as inter-category ambiguity and intra-category variation have been holding back their performance. 
To overcome such issues, gathering semantic information has been shown to be a promising source of information towards a more complete and discriminative feature representation of indoor scenes. Therefore, the work described in this paper uses both semantic information, obtained from object detection, and semantic segmentation techniques. While object detection techniques provide the 2D location of objects allowing to obtain spatial distributions between objects, semantic segmentation techniques provide pixel-level information that allows to obtain, at a pixel-level, a spatial distribution and shape-related features of the segmentation categories.
Hence, a novel approach that uses a semantic segmentation mask to provide Hu-moments-based segmentation categories' shape characterization, designated by Segmentation-based Hu-Moments Features (SHMFs), is proposed. Moreover, a three-main-branch network, designated by GOS$^2$F$^2$App, that exploits deep-learning-based global features, object-based features, and semantic segmentation-based features is also proposed. GOS$^2$F$^2$App was evaluated in two indoor scene benchmark datasets: SUN RGB-D and NYU Depth V2, where, to the best of our knowledge, state-of-the-art results were achieved on both datasets, which present evidences of the effectiveness of the proposed approach.
\end{abstract}


\section{Introduction}
\label{sec:intro}

Indoor scene classification is an open and challenging computer vision task that assigns a scene category to the input data, widely applied \cite{survey_2020} in autonomous mobile robots, and applications of human-computer interaction. 

A scene category is represented by the semantic cues, which are composed of the objects scattered around the scene and their relationships \cite{Montoro_2021,mapnet_2019}. However, obtaining a distinctive scene feature representation has become a challenging problem due to the rise of two major issues in the indoor scene classification task: 
intra-category variation and inter-category ambiguity. The intra-category variation is caused by having, for the same scene category, multiple points of view and various objects with diverse layouts, which make the extraction of common patterns more complex \cite{df2net_2018,gsf2appV2_IROS21}. The inter-category ambiguity between different scene categories often occurs when scene categories share similar objects' appearance and distribution. %\cite{cheng_sdo_2018}. 
This ambiguity blurs the inter-scene boundaries \cite{lopezcifuentes_SceneSeg_2020PR}, making it difficult to obtain a distinctive scene feature representation that leads to poor classification performances.
In an attempt to overcome the aforementioned issues, researchers have exploited different approaches such as global features \cite{df2net_2018}, local features \cite{lopezcifuentes_SceneSeg_2020PR}, and mid-level visual representations \cite{caglayan_2022}.

%In the past few years, with the dramatic growth in performance of deep learning approaches, Convolutional Neural Networks (CNNs) in particular, have led to significant performance enhancements on computer vision tasks \cite{xiong_2019,gs2f2app_ICRA23}. CNN-based methods can extract complex global features and find, in an effective way, patterns between images from the same categories, which also led to performance boosts in scene classification tasks. %However, when applied to the indoor scene classification task, extracted global features are too spatially rigid to encode the invariance of semantic information, being unable to represent the scene and overcome the aforementioned limitations \cite{mapnet_2019,Xiong_2021,xiong_2019}.
%Therefore, aiming to obtain a better scene representation, new approaches and sources of information about the scene are required, such as representing scene categories based on the semantic information \cite{mapnet_2019,Cai_2019}.

In recent years, the huge performance growth of deep learning approaches, in particular Convolutional Neural Networks (CNNs), has led to significant performance improvements in computer vision tasks \cite{xiong_2019,gs2f2app_ICRA23}. However, when applied to the indoor scene classification task, extracted global features are too spatially rigid to encode the invariance of semantic information, being unable to represent the scene and overcome the aforementioned limitations \cite{mapnet_2019,Xiong_2021}. Therefore, aiming to obtain a better scene representation, new approaches and sources of information about the scene, such as the representation of scene categories based on semantic information, \cite{mapnet_2019,Cai_2019}, are required.

To recognize a scene category, humans need to identify the objects available in the scene and perceive the context by considering correlations between identified objects \cite{mapnet_2019,df2net_2018}. To complement the deep-learning-based global features, the works proposed in \cite{Cai_2019,song_oor_2020} used semantic information to reduce the variation among intra-category scenes and decrease the ambiguity between scene categories. To obtain semantic features, the methods proposed in \cite{gsf2appV2_IROS21, lopezcifuentes_SceneSeg_2020PR} have resorted to object detection techniques or semantic segmentation approaches. The majority of object detectors provide, in the form of 2D bounding boxes, the location and bounding box dimensions of the recognized objects and their categories. Hence, object bounding boxes can be used to obtain spatial distributions of objects from the same category and from objects from different categories. On the other hand, object detectors do not provide enough information regarding an object's shape, which may be important to disentangle scene predictions. 
Semantic segmentation approaches provide classification at the pixel level, i.e., at each pixel is assigned a segmentation-category that can be an object category, wall, floor, or another type of category, allowing to obtain a well-defined segmentation-category's shape estimation. 
Thus, having in view a better semantic representation of the scene, semantic information gathered by object detectors can be complemented with semantic information provided by semantic segmentation approaches.
Furthermore, while semantic segmentation approaches provide a way to estimate segmentation-categories' shapes, they are not able to characterize them, leading to the need for additional features to perform their characterization. Hu-moments \cite{HuMoments_1962} are an image descriptor widely applied to characterize image-based objects' shapes \cite{HM_aplications_PR_2019}. They are composed of seven moments (features) based on the image's intensity distribution and are invariant to translation, rotation, scale, and reflection transformations. Therefore, the use of Hu-moments applied to the segmentation-categories' shapes may allow to obtain a unique semantic representation of the scene that may aid in overcoming the aforedescribed issues on the scene classification task, by decreasing the ambiguity between scene categories and reducing the variation between the same scene category. Hence, having such representation complemented with additional semantic information provided by object detectors and/or other semantic segmentation-based features can lead to a better semantic representation of the scene, which will improve the overall scene classification performance.

\begin{figure*}[!tb]
	\centering
	\includegraphics[trim={0cm 0.15cm 0cm 0cm}, clip, width=0.95\linewidth]{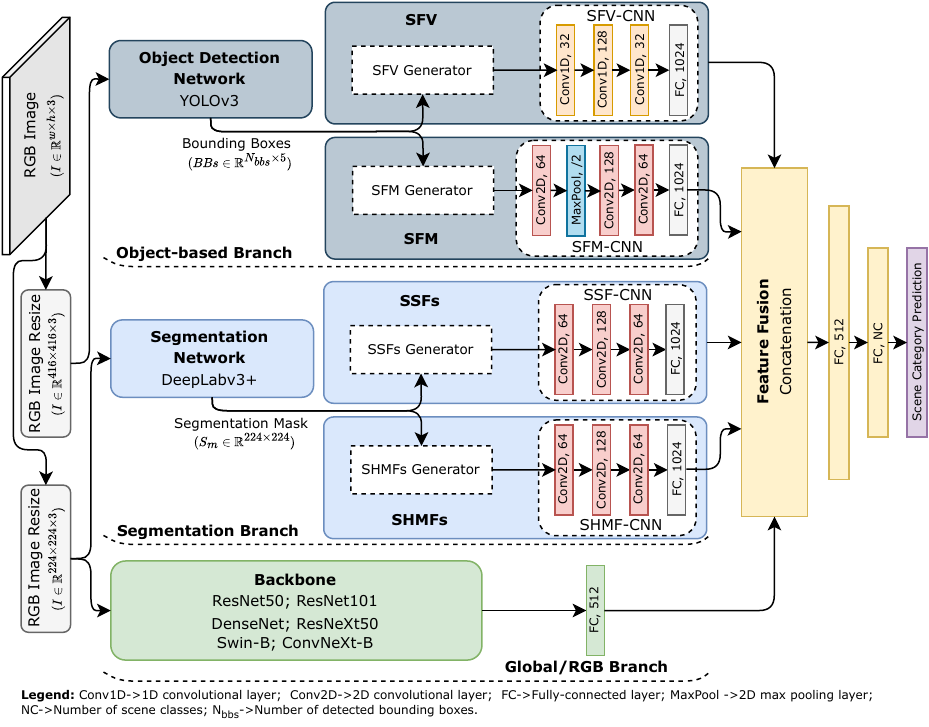}
	\caption{Overview of the proposed GOS$^2$F$^2$App, which consists of three main branches. The object-based branch (top branch) generates object bounding boxes, from which the SFV and SFM object-based features are extracted. These features are further exploited by the SVF-CNN and SFM-CNN, respectively. In the segmentation branch (middle branch), a semantic segmentation mask is generated, from which the SHFMs and SSFs segmentation-based features are extracted. These features are later exploited by the SHFM-CNN and SSF-CNN, respectively. The global branch (bottom branch) employs a state-of-the-art deep-learning-based feature extraction approach to learn global features. All the output features converge to a feature fusion stage to obtain a scene category prediction.}
	\label{fig:gos2f2app_overview}	
\end{figure*}

In this paper, aiming to obtain a more discriminative feature representation of indoor scenes, a novel approach that uses semantic segmentation masks to generate Segmentation-based Hu-Moments Features (SHMFs) is proposed. In detail, SHMFs encode a segmentation-categories' shape characterization obtained from adapted Hu-moments \cite{HuMoments_1962}.
Moreover, a three-main-branch deep-learning-based Global, Object-based, and Semantic Segmentation Feature Fusion Approach (GOS$^2$F$^2$App), as shown in Fig. \ref{fig:gos2f2app_overview}, is also proposed. It exploits deep-learning-based global features from RGB images using a state-of-the-art network, object-based features generated from object bounding boxes provided by YOLOv3 \cite{yolov3_2018}, and semantic segmentation-based features generated from semantic segmentation masks provided by DeepLabv3+ \cite{DeepLabv3_2018}. Object-based features are composed of the Semantic Feature Vector (SFV) \cite{gsf2app_ICARSC20} and the Semantic Feature Matrix (SFM) \cite{gsf2appV2_IROS21}, which encode the objects' occurrences and distance relationships between object categories, respectively. On the other hand, semantic segmentation-based features are composed of the proposed SHMFs and the Segmentation-based Semantic features (SSFs) \cite{gs2f2app_ICRA23}, which encode a 2D spatial layout of the segmentation-categories over the scene. All branches' outputs undergo a feature fusion stage, where all features are concatenated for a scene category prediction.

The GOS$^2$F$^2$App was evaluated on two popular indoor scene benchmark datasets, the SUN RGB-D \cite{sun_dataset} and the NYU Depth Dataset V2 \cite{nyu_dataset}, being able to achieve state-of-the-art results on both datasets.

%As aforementioned, this work uses the YOLOv3 \cite{yolov3_2018} object detector network and the DeepLabv3+ \cite{DeepLabv3_2018} semantic segmentation network, from which semantic features are extracted. The performance of both networks may influence the GOS$^2$F$^2$App overall performance, since a poor object detection performance, as well as a poor segmentation performance could lead to an unreliable feature representation, which in turn will lead to a poor scene classification performance. The generalization capabilities of both object detector and segmentation network are not included in the scope of this work, being the main focus on assessing the performance of the proposed semantic feature representation in the GOS$^2$F$^2$App. However, in line with the main focus of this work, similar to \cite{gs2f2app_ICRA23}, three semantic segmentation models with different overall performances are used to evaluate and assess the stability of the proposed GOS$^2$F$^2$App.

In summary, the main contributions of this work are as follows:

\begin{itemize}
    \item A novel semantic segmentation-based approach that extracts SHMFs, which encode a Hu-moments-based characterization of segmentation-categories' shapes;
    \item GOS$^2$F$^2$App: a tree-main-branch network that seeks to overcome the inter-category ambiguity and intra-category variation issues by exploiting global, object-based, and semantic segmentation-based features.  %global features, object-based features, and semantic segmentation-based features;
    \item A comprehensive study is provided to assess the influence of semantic features in the indoor scene classification task, the influence of using semantic segmentation models with different overall performances, and the influence of different deep-learning-based state-of-the-art networks on extracting global features;
    \item The GOS$^2$F$^2$App was evaluated on the SUN RGB-D dataset \cite{sun_dataset} and NYU Depth Dataset V2 \cite{nyu_dataset}, attaining, to the best of our knowledge, state-of-the-art performance on both datasets.
\end{itemize}

%%%%%%%%%%%%%%%%%%%%%%%%%%%%%%%%%%%%%%%%%%%%%%%%%%%%%%%%%%%%%%%%%%%%%%%%%%%%%%%%
\section{Related Work}
\label{sec:related_work}

The emergence of deep learning techniques has sparked researchers' interest in indoor scene classification, leading to the development of various methodologies and approaches to solve this challenging problem \cite{survey_2020}.
Despite the variety of techniques applied in each work, most of them follow a three-stage pipeline consisting of feature extraction, transformation and aggregation, and scene category prediction \cite{survey_2020,mapnet_2019}. Large-scale datasets like the object-centric ImageNet dataset \cite{imagenet_dataset} and scene-centric Places dataset \cite{places_dataset} have been proposed to train deep learning models. However, when pre-trained models were applied to indoor scene datasets, the extracted features remained too coarse to represent an indoor scene \cite{mapnet_2019,df2net_2018}. The variation and ambiguity found in indoor scenes make full-image deep learning-based global features insufficient to represent them \cite{xiong_2019}. To overcome this issue, researchers have focused on developing new feature representations, mainly multi-modal and semantic features.

\subsection{Multi-Modal Features}
With the need of gathering new sources of scene information, researchers have been developing multi-modality approaches, with a greater focus on RGB and depth modalities \cite{mapnet_2019,caglayan_2022, du_pyramid_2021}. 
Since depth data can be used to obtain semantic cues regarding objects' shape, approaches exploiting how to effectively extract RGB-D features and how to combine both modalities have been proposed. 
%Wang \textit{et al.} \cite{wang_2016} extracted global and local features from different modalities, which were combined using a Fisher vector encoding approach to learn component-aware representations. However, such approaches, where the main contribution was on aggregating feature-level relationships are enforcing local features in multi-modalities to follow the same relationships in fusion, which are not appropriate for indoor scenes \cite{mapnet_2019}.
%Therefore, towards a better feature representation, multi-branch networks were proposed, aiming to individually extract features from each modality  \cite{mapnet_2019,df2net_2018, Xiong_2021,xiong_2019}. 
Li \textit{et al.} \cite{df2net_2018} proposed a two-stage learning approach that, in the first stage, learns feature representations on each modality, while in the second stage learns correlative features between both modalities. Li \textit{et al.} \cite{mapnet_2019} improved \cite{df2net_2018} by integrating RGB-D semantic cues, obtained through a region proposal technique, which are further exploited by attentive pooling blocks. However, approaches that used semantic cues generated by region proposal techniques are not able to generate enough discriminative features to overcome limitations such as intra-category variation and inter-category similarity. 
Hence, approaches that exploited global, local, and multi-modality correlation features were proposed in \cite{xiong_2019,Xiong_2021,xiong_MSN_2020}. Despite the performance enhancement presented by multi-branch networks, Du \textit{et al.} \cite{du_pyramid_2021} argues that using modality-specific networks to learn representations for each modality fails to learn important correlations between modalities. Then, Du \textit{et al.} \cite{du_pyramid_2021} proposed a cross-modal pyramid translation framework with shared features between modalities. On the other hand, Caglayan \textit{et al.} \cite{caglayan_2022} extracted CNN-based RGB-D features at multiple levels, which were mapped into high-level representations using a fully randomized structure of recursive neural networks. Such approaches can learn, in more detail, multi-modal correlated features, which allows to learn more discriminative features of the scene. 
%However, despite the semantic scene information that can be gathered through depth data, the aforementioned works used semantic cues based on shapes extracted from depth data and do not take into consideration the object categories and their distribution over the scene, which can play an important role in disentangling the scene predictions. 
Aforementioned works used depth data to gather semantic scene information, but they only relied on shape cues and did not consider object categories and their distribution in the scene, which can play an important role in disentangling the scene predictions.

\subsection{Semantic Features}
Aiming for a more interpretable and discriminative feature representation of the scene, researchers have been exploiting, in addition to CNN-based global features, semantic-based features focused on the objects recognized on the indoor scene \cite{gsf2appV2_IROS21, song_oor_2020, gsf2app_ICARSC20}. 
%In the literature, indoor scene classification pipelines that exploited semantic-based features can be unfolded in two major categories: i) image-based local features; ii) object-based features.
In the literature, such approaches can be unfolded in two major categories: i) image-based local features; ii) object-based features.

Approaches that fall into the image-based local features category focus on detecting regions of interest (RoIs) and learning correlations among them. Laranjeira \textit{et al.} \cite{Laranjeira_2019} proposed a bidirectional long short-term memory network that correlates features based on detected RoIs to learn underlying scene layouts. 
%Xie \textit{et al.} \cite{Xie_Hierarchical_2020} and 
Cai \& Shao \cite{Cai_2019} used image-based patches based on detected RoIs to force CNNs to learn local-based visual concepts and correlation patterns between local features and indoor scenes. Such approaches are not invariant to the scene layout, since spatial-related information is not taken into consideration \cite{chen_2020}. Therefore, to learn and encode the scene's spatial layouts, works using CNN-based graph networks were proposed \cite{chen_2020,Yuan_2019}. Yuan \textit{et al.} \cite{Yuan_2019} selected object-based RoIs as graph nodes and Chen \textit{et al.} \cite{chen_2020} proposed a layout graph network with two sub-graphs to capture spatial and semantic similarity relationships between object-based RoIs, which were used as graph nodes. Despite the performance improvement achieved by such approaches, their feature representation remains too coarse to classify indoor scenes in scenarios where a single scene category layout varies too much, and scene categories present a high degree of similarity. Also, since RoIs methods output 2D bounding box coordinates without any object category prediction, approaches relying on RoIs as a local feature generator are unable to obtain a semantic categorization and their crucial spatial correlations, which are key to overcoming the aforementioned issues.

On the other hand, approaches using object-based features have been proposed in \cite{gsf2appV2_IROS21,song_oor_2020, George_2016}. To exploit object relationships, these approaches use recognized objects in the scene, as a source of semantic information. George \textit{et al.} \cite{George_2016} modeled the occurrence patterns of objects in scenes. Then, such occurrences were transformed into scene probabilities. Seong \textit{et al.} \cite{Seong_fosnet_2020} proposed a two-stream network that exploits CNN-based global and CNN-based object features. Pereira \textit{et al.} \cite{gsf2app_ICARSC20} proposed a similar approach, being the object features represented by a semantic vector containing the occurrences of objects. 
%Sitaula \textit{et al.} \cite{Sitaula_2019} developed an object pattern dictionary to map the occurrences of objects to indoor scenes. 
However, none of the previous object-based feature approaches incorporate object-based spatial information, leading to misclassifications when similar objects appear in different scene categories. 
%Hence, Zeng \textit{et al.} \cite{Zeng_2020} extracted, in addition to global and object-based features, scene attributes in local regions of the scene. 
%Cheng \textit{et al.} \cite{cheng_sdo_2018} exploited the correlations of object configurations across each scene by capturing statistical information on the co-occurrence patterns of objects. 
Hence, Song \textit{et al.} \cite{song_2017} proposed an object-to-object correlation method to represent spatial layouts of scene categories, which was later improved in \cite{song_oor_2020} by exploiting the frequency and sequential representations of object-to-object relationships. Pereira \textit{et al.} \cite{gsf2appV2_IROS21} proposed an inter-object distance relationship approach that encodes how close or apart object classes are over the scene. The aforementioned works have achieved meaningful and discriminative feature representations of scene categories, leading to a significant performance improvement in the indoor scene classification task. 
However, approaches that use, as a source of semantic information, object detector techniques to recognize the objects available in the scene, cannot extract accurate information about the whole scene (e.g. floor, walls) or pixel-level object's shape that may be crucial to overcoming the remaining indoor scene classification issues.

To obtain more meaningful object correlation representations, approaches using semantic segmentation masks as a source of semantic information have been proposed in \cite{lopezcifuentes_SceneSeg_2020PR, gs2f2app_ICRA23, Zhou_iros_2021}. 
%Herranz-Perdiguero \textit{et al.} \cite{Perdiguero_SceneSeg_2018Iros} proposed a histogram-like approach that represents the number of pixels of each semantic category. 
%Ahmed \textit{et al.} \cite{Ahmed_SceneSeg_2020Sensors} proposed segmentation-based descriptors for multiple object categorization. 
Zhou \textit{et al.} \cite{Zhou_iros_2021} proposed a Bayesian object relationship model that analyzes the object co-occurrences and pairwise object relationships. López-Cifuentes \textit{et al.} \cite{lopezcifuentes_SceneSeg_2020PR} 
%and Miao \textit{et al.} \cite{Miao_iros_2021} 
used segmentation masks to extract object relationships, which were combined with CNN-based global features. Pereira \textit{et al.} \cite{gs2f2app_ICRA23}, proposed a segmentation-based 2D spatial layout of the object categories across the scene. Such approaches that rely on semantic segmentation masks can gather meaningful and promising semantic information about the scene. However, when multiple objects of the same category are too close, semantic segmentation masks will not be able to distinguish and represent them as multiple objects, instead, they will consider them as one single object, which can negatively impact the generated feature representation of the scene. Therefore, motivated by the positive aspects that object-based and semantic-based features offer to obtain a discriminative feature representation of the scene, the herein proposed approach exploits semantic features provided by both sources of semantic information.

%%%%%%%%%%%%%%%%%%%%%%%%%%%%%%%%%%%%%%%%%%%%%%%%%%%%%%%%%%%%%%%%%%%%%%%%%%%%%%%%
\section{Methodology}
\label{sec:methodology}

Figure \ref{fig:gos2f2app_overview} presents an overview of the proposed GOS$^2$F$^2$App, which is composed of three main branches that exploit deep-learning-based global features, segmentation-based features, and object-based features. All branches' output features undergo a feature fusion stage, where additional feature correlations are obtained for scene category prediction.

\subsection{Segmentation Branch}

A semantic segmentation mask provides a pixel-level classification of an indoor scene image that allows the extraction of unique semantic information about the object categories, furniture, floors, and walls. To obtain a descriptive semantic feature representation, two different types of segmentation-based features extracted from semantic segmentation masks are adopted: the segmentation-based Hu-moments features (SHMFs) proposed in this paper, and the segmentation-based semantic features (SSFs) proposed in \cite{gs2f2app_ICRA23}. To obtain semantic segmentation masks from RGB images, the DeepLabv3+ \cite{DeepLabv3_2018} encoder-decoder-based semantic segmentation network is used.

\subsubsection{Segmentation-based Hu-Moments Features}
%https://learnopencv.com/shape-matching-using-hu-moments-c-python/
%https://cvexplained.wordpress.com/2020/07/21/10-4-hu-moments/
A semantic segmentation mask provides, at a pixel-level, meaningful information about the segmentation-categories, which allows to obtain a well-defined shape of each category. Therefore, to take advantage of the categories' shapes, the seven Hu-moments \cite{HuMoments_1962} are used. The Hu-moments are based on the image normalized central moments, which are calculated using the centroids of the image moments, making them invariant to translation, scale, rotation and reflection transformations. Due to their properties, they were widely applied, before the advent of deep learning methods, as an image descriptor to quantify an object's shape \cite{HuMoments_1962,HM_aplications_PR_2019}. In this work, the seven Hu-moments are adapted to be applied in each segmentation-category available in the generated semantic segmentation mask.

Let $L$ be the total number of segmentation-categories, $S_m$ a semantic segmentation mask with size $w\times h$, $n\in [1, L]$ an index identifying an arbitrary segmentation-category, and $S_m(i,j) \in [1, L]$ a pixel value representing an index corresponding to a segmentation-category at the indexes $i = [1,...,h]$, and $j=[1,..,w]$.
Hence, per segmentation-category, the raw moment ($M_{pq}$) of order $(p+q)$, which represents the shape of the segmentation-category, is calculated as follows:

\small
\begin{equation}
    \label{eq:moments}
    M_{pq}(n) = \sum_{j=1}^{w}{\sum_{i=1}^{h}{}} j^p i^q f_C(i,j,n),
\end{equation}
with
\begin{equation}
    f_C(i,j,n) = \left\{\begin{matrix} 
        1 & \textup{if } S_m(i,j)= n\\ 
        0 & \textup{otherwise }
        \end{matrix}\right.
        ,
\end{equation}
\normalsize
where $p,q \in \mathbb{Z}^{0+}$. Then, to obtain moments invariant for translation transformations, designated by central moments $(\mu_{pq})$, the raw moment's centroid $(\bar{C_i},\bar{C_j})$ is required, which is obtained as follows:

\small
\begin{equation}
    \label{eq:centroid}
    \bar{C_j}(n) = \frac{M_{10}(n)}{M_{00}(n)} \ \ \ \ \ \ \bar{C_i}(n) = \frac{M_{01}(n)}{M_{00}(n)}
\end{equation} \normalsize
and the central moments are calculated as follows:
\small
\begin{equation}
    \label{eq:central_moments}
    \mu_{pq}(n) = \sum_{j=1}^{w}{\sum_{i=1}^{h}{}} (j-\bar{C_j}(n))^p\ (i-\bar{C_i}(n))^q\  f_C(i,j,n)
\end{equation} \normalsize
To obtain a scale invariant transformation, the normalized central moments ($\eta_{pq}$) are obtained as follows:
\small
\begin{equation}
    \label{eq:norm_central_moments}
    \eta_{pq}(n) = \frac{\mu_{pq}(n)}{\mu_{00}^r(n)}, 
\end{equation} \normalsize
being $r = (p+q)/2+1$.
Then, according to Hu \cite{HuMoments_1962}, combining the normalized central moments using the 2nd and 3rd order moments, seven invariant moments can be obtained  as follows:
\small \begin{equation}
    \label{eq:h1}
   h_1(n) = \eta_{20}(n) + \eta_{02}(n)
\end{equation} \normalsize
\small \begin{equation}
    \label{eq:h2}
   h_2(n) = (\eta_{20}(n) - \eta_{02}(n))^2 + 4\eta_{11}^2(n)
\end{equation} \normalsize
\small \begin{equation} 
    \label{eq:h3}
   h_3(n) = (\eta_{30}(n) - 3\eta_{12}(n))^2 + (3\eta_{21}(n)-\eta_{03}(n))^2
\end{equation} \normalsize
\small \begin{equation} 
    \label{eq:h4}
   h_4(n) = (\eta_{30}(n) + \eta_{12}(n))^2 + (\eta_{21}(n)+\eta_{03}(n))^2
\end{equation} \normalsize
\small %\begin{equation}
    \begin{multline}
    \label{eq:h5}
    h_5(n) = (\eta_{30}(n)-3\eta_{12}(n))(\eta_{30}(n)+\eta_{12}(n))[(\eta_{30}(n)+\eta_{12}(n))^2 \\  -3(\eta_{21}(n)+\eta_{03}(n))^2] + (3\eta_{21}(n)-\eta_{03}(n))\\ [3(\eta_{30}(n)+\eta_{12}(n))^2-(\eta_{21}(n)+\eta_{03}(n))^2]
    \end{multline}
%\end{equation} \normalsize
\small \begin{multline}
    \label{eq:h6}
    %\begin{split}
    h_6(n) = (\eta_{20}(n)-\eta_{02}(n))[(\eta_{30}(n)+\eta_{12}(n))^2  -(\eta_{21}(n)\\+\eta_{03}(n))^2] +4\eta_{11}(n)(\eta_{30}(n)+\eta_{12}(n))(\eta_{21}(n)+\eta_{03}(n))
    %\end{split}
\end{multline} \normalsize
\small \begin{multline}
    \label{eq:h7}
    %\begin{split}
    h_7(n) = (3\eta_{21}(n)-\eta_{03}(n))(\eta_{30}(n)+\eta_{12}(n))[(\eta_{30}(n)+\eta_{12}(n))^2 \\ -3(\eta_{21}(n)+\eta_{03}(n))^2]  - (\eta_{30}(n)-3\eta_{12}(n))(\eta_{21}(n)+\eta_{03}(n))\\ [3(\eta_{30}(n)+\eta_{12}(n))^2-(\eta_{21}(n)+\eta_{03}(n))^2]
    %\end{split}
\end{multline} \normalsize
In more detail, the first six moments ($h_1(n), \cdots, h_6(n) $) are invariant to translation, scale, rotation, and reflection transformations. To allow distinguishing mirror object's shapes, the last moment ($h_7(n)$) is only invariant to translation, scale and rotation transformations, since the application of a reflection transformation changes its sign. Moreover, the generated  Hu-moments values can reach different orders of magnitude, which make them not comparable. To overcome such issue, the Hu-moments values are re-scaled, through a log transformation, as expressed:

\small \begin{equation}
    h_k'(n) = -sign(h_k(n))\times\log|h_k(n)|,\  k \in [1,7]
\end{equation} \normalsize
All obtained Hu Moments, per segmentation-category, are rearranged under a matrix form, $SHMFs \in \mathbb{R}^{l\times 7}$, represented as follows:

%\small \begin{equation}
%\label{matrix:shmf}
%SHMFs = 
%    \begin{bmatrix}
%        h_1'(1) & h_2'(1) & h_3'(1) & h_4'(1) & h_5'(1) & h_6'(1) & h_7'(1) \\
%        h_1'(2) & h_2'(2) & h_3'(2) & h_4'(2) & h_5'(2) & h_6'(2) & h_7'(2) \\
%        \vdots & \vdots & \vdots & \vdots & \vdots & \vdots & \vdots \\
%        h_1'(l) & h_2'(l) & h_3'(l) & h_4'(l) & h_5'(l) & h_6'(l) & h_7'(l) \\
%    \end{bmatrix}  
%\end{equation} \normalsize

\small \begin{equation}
\label{matrix:shmf}
SHMFs = 
    \begin{bmatrix}
        h_1'(1) & h_2'(1) & \cdots & h_6'(1) & h_7'(1) \\
        h_1'(2) & h_2'(2) & \cdots & h_6'(2) & h_7'(2) \\
        \vdots & \vdots & \vdots & \vdots & \vdots \\
        h_1'(L) & h_2'(L) & \cdots & h_6'(L) & h_7'(L) \\
    \end{bmatrix} 
    \in \mathbb{R}^{L\times 7},
\end{equation} \normalsize

\begin{figure}[!tb]
    \centering
    \includegraphics[width=1.0\linewidth]{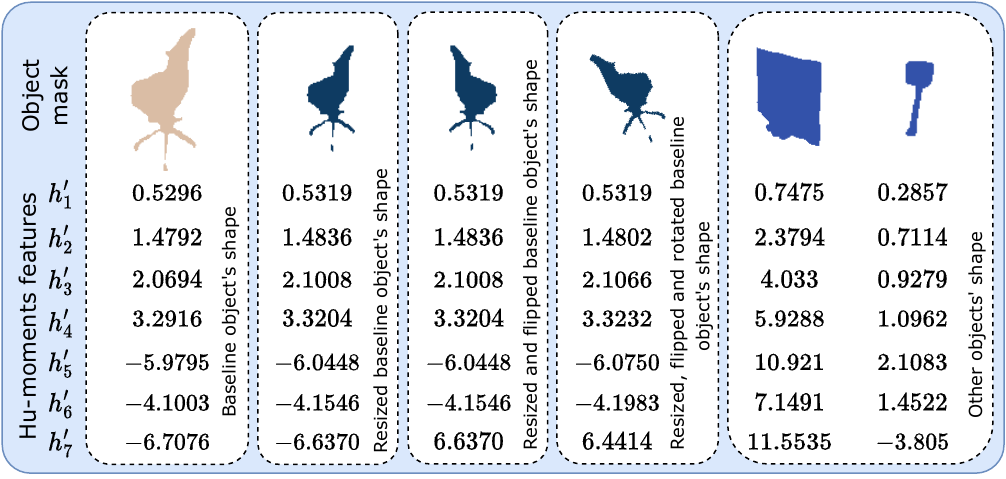}
    \caption{Example of Hu-moments features obtained in different objects' shape conditions.}
    \label{fig:huMoments}	
\end{figure}

An example of the invariant and non-invariant Hu-moments is shown in Fig. \ref{fig:huMoments}, which presents re-scaled Hu-moments values obtained from segmentation-categories' shape. First, Hu-moments values using four chairs' shapes are presented, being the last three chairs' shapes obtained by applying scale, reflection, and rotation transformations in the first chair's shape. It can be observed that regardless of the transformation applied, the resulting Hu-moments values are similar for all moments, being observed a change of signal in the last moment value when a reflection transformation was applied. Then, by comparing Hu-moments values obtained for the chairs' shapes with Hu-moments values obtained for other segmentation-categories (tv and lamp), it is possible to observe significant differences between them, demonstrating the effectiveness of the Hu-moments values in obtaining a discriminate characterization of each segmentation-category's shape. Based on the aforementioned, the Hu-moments-based features can provide meaningful and descriptive information resorting to the semantic categories available in indoor scenes, which can allow to improve indoor scene feature representations.

%\subsubsection*{SHMF-CNN Architecture}
To learn additional feature correlations from the proposed SHMFs, a CNN-based architecture, SHMF-CNN (as shown in Fig. \ref{fig:gos2f2app_overview}), is used. It is composed of three 2D convolutional layers (with 64, 128, and 64 output channels, respectively) followed by a fully-connected (FC) layer. Each convolutional layer uses a kernel size of 3, a stride value of 1, and a padding value of 1. All convolutional and FC layers use the ReLU as the activation function.

\subsubsection{Segmentation-based Semantic features} 

Despite the discriminative feature representation provided by the proposed SHMFs, which encodes a Hu-moments-based characterization of segmentation-categories' shape, they do not take into consideration a spatial distribution of the segmentation-categories. To obtain a more complete segmentation-based feature representation, the SSFs proposed in \cite{gs2f2app_ICRA23} are used in addition to the proposed SHMFs. The SSFs represent how much a specific segmentation-category is present in the scene and how segmentation-categories are distributed over the indoor scene. They contain the following data related to each segmentation-category: 1) the pixel count ($P_C$) that represents how much a specific segmentation-category is presented; ii) the 2D average position ($I_{\mu_x}$, $I_{\mu_y}$) which represents where each segmentation-category is centered in the scene; iii) the 2D standard deviation values ($I_{\sigma_x}$, $I_{\sigma_y}$) based on each 2D pixel position, which represent how close or apart the same segmentation-category is across the scene. SSFs are expressed as follows:

\small \begin{equation}
\label{matrix:ssf}
SSFs = 
    \begin{bmatrix}
        P_C'(1) & I_{\mu_x}'(1) & I_{\mu_y}'(1) & I_{\sigma_x}'(1) &             I_{\sigma_y}'(1) \\
        P_C'(2) & I_{\mu_x}'(2) & I_{\mu_y}'(2) & I_{\sigma_x}'(2) & I_{\sigma_y}'(2) \\
        \vdots & \vdots & \vdots & \vdots & \vdots \\
        P_C'(L) & I_{\mu_x}'(L) & I_{\mu_y}'(L) & I_{\sigma_x}'(L) & I_{\sigma_y}'(L) \\
    \end{bmatrix}  
    \in \mathbb{R}^{L\times 5}.
\end{equation} \normalsize
where each element of the SSFs is calculated as follows:

\noindent\begin{minipage}{.6\linewidth}
    \small \begin{equation}
    \label{eq:pixel_number}
        P_C(n) = \sum_{i=1}^{h}{\sum_{j=1}^{w}{}} f_C(i,j,n)
    \end{equation} \normalsize
\end{minipage}%
\begin{minipage}{.4\linewidth}
    \small \begin{equation}
    \label{eq:norm_pixel_number}
        P_C'(n) = \frac{P_C(n)}{h\times w}  
    \end{equation} \normalsize
\end{minipage}

\small \begin{equation}
    \label{eq:I_mu_x}
    I_{\mu_x}(n) = \frac{\sum_{i=1}^{h}{\sum_{j=1}^{w}{}} j \times f_C(i,j,n) }{P_C(n)}
\end{equation} \normalsize
 
\small \begin{equation} 
    \label{eq:I_mu_y}
    I_{\mu_y}(n) = \frac{\sum_{i=1}^{h}{\sum_{j=1}^{w}{}} i \times f_C(i,j,n) }{P_C(n)}
\end{equation} \normalsize

\noindent\begin{minipage}{.5\linewidth}
    \small \begin{equation}
    \label{eq:norm_I_mu_x}
        I_{\mu_x}'(n) = \frac{1}{w} I_{\mu_x} (n)
    \end{equation} \normalsize
\end{minipage}%
\begin{minipage}{.5\linewidth}
    \small \begin{equation}
    \label{eq:norm_I_mu_y}
        I_{\mu_y}'(n) = \frac{1}{h} I_{\mu_y} (n)
    \end{equation} \normalsize
\end{minipage}

\small \begin{equation}
    \label{eq:I_sigma_x}
    I_{\sigma_x}(n) = \sqrt{\frac{\sum_{i=1}^{h}{\sum_{j=1}^{w}{}} (j-I_{\mu_x}(n))^2 \times f_C(i,j,n)}{P_C(n)}} 
\end{equation} \normalsize
\small \begin{equation}
    \label{eq:I_sigma_y}
    I_{\sigma_y}(n) = \sqrt{\frac{\sum_{i=1}^{h}{\sum_{j=1}^{w}{}} (i-I_{\mu_y}(n))^2 \times f_C(i,j,n)}{P_C(n)}} 
\end{equation} \normalsize

\noindent\begin{minipage}{.5\linewidth}
    \small \begin{equation}
        \label{eq:norm_I_sigma_x}
        I_{\sigma_x}'(n) = \frac{1}{w} I_{\sigma_x}(n)
    \end{equation} \normalsize
\end{minipage}%
\begin{minipage}{.5\linewidth}
    \small \begin{equation}
        \label{eq:norm_I_sigma_y}
        I_{\sigma_y}'(n) = \frac{1}{h} I_{\sigma_y}(n)
    \end{equation} \normalsize
\end{minipage}
being $P_C'$ a normalized value of $P_C$, ($I_{\mu_x}'$, $I_{\mu_y}'$) normalized values of ($I_{\mu_x}$, $I_{\mu_y}$), and ($I_{\sigma_x}'$, $I_{\sigma_y}'$) normalized values of ($I_{\sigma_x}$, $I_{\sigma_y}$).

%\subsubsection*{SSF-CNN Architecture}
To exploit additional feature correlations from the SSF, a CNN-based architecture, SSF-CNN (see Fig. \ref{fig:gos2f2app_overview}), similar to the SHMF-CNN, is used.

\subsection{Object-based Branch}
%Similar to the features extracted from semantic segmentation masks, 2D object-related features, obtained from the output of an object detection module, can also provide unique and relevant semantic information about the indoor scene, by representing the objects and their inter-relationships. Furthermore, combining object-segmentation features with object-related features allow for a more complete and meaningful semantic feature representation of the scene. 

Segmentation-based features provide rich and meaningful semantic information about the whole scene image. However, such features fail to represent the occurrences that each object category has in the scene as well as object occurrence-based correlated features. Therefore, object-based features generated from the output of an object detection module can provide unique and relevant semantic information that can complement the segmentation-based features, hopefully contributing to obtain a more complete and meaningful semantic feature representation of the scene. To detect and classify the objects available in the scene provided by means of RGB images, the YOLOv3 \cite{yolov3_2018} object detector network is used, with a COCO dataset \cite{coco_data} pre-trained model that is able to detect eighty different object categories ($N=80$). Thus, using the objects' location and classification provided by the YOLOv3 network, two different types of object-based features are generated: semantic feature vector (SFV) \cite{gsf2app_ICARSC20} and semantic feature matrix (SFM) \cite{gsf2appV2_IROS21}.

\subsubsection{Semantic Feature Vector}
%Object categories can be shared between different indoor scene categories, meaning that recognizing that a determined object category is in the scene may not be enough to obtain an unique semantic representation. 
%Therefore, the SFV, proposed in \cite{gsf2app_ICARSC20} is used. The SFV encodes the number of occurrences that each object category is recognized in a scene image ($O_{i=1:N}$) and is represented as follows:

Some scene categories can be often characterized by the objects recognized on them. Such representation can be obtained by encoding the number of occurrences that each object category has in each scene category. Hence, the SFV proposed in \cite{gsf2app_ICARSC20}, encodes the number of occurrences that each object category ($O_{i=1:N}$) is recognized in a scene image, which is represented as follows:

\small \begin{equation}
SFV = 
\begin{bmatrix}
    O_{1} & O_{2} & \cdots & O_{N}
    \end{bmatrix}  
\end{equation} \normalsize

%\subsubsection{SFV-CNN Architecture}
The generated SFV are further exploited by a CNN-based architecture \cite{gsf2appV2_IROS21}, SFV-CNN. The network is composed of three 1D convolutional layers (with 32,128,32 output channels, respectively) followed by a FC layer, as shown in Fig. \ref{fig:gos2f2app_overview}. Convolutional layers are composed of a kernel size of 3, a stride value of 1, and a padding value of 1. The ReLU activation function is applied in both convolutional and FC layers.

\subsubsection{Semantic Feature Matrix}

Object categories can be shared between different indoor scene categories, meaning that recognizing a set of object categories in a scene is not enough to obtain a unique semantic representation of the scene. In an attempt to obtain a more discriminative object category-based scene representation, object-based spatial features encoding the distribution of the objects over the scene, are required. Hence, the SFM proposed in \cite{gsf2appV2_IROS21}, containing inter-object distance relationships ($S=\{b_{(1,1,1:K)},\dots,b_{(N,N,1:K)}\}$ in a histogram-like approach is used. Such features represent how close or apart objects belonging to two object categories are, and are expressed as follows:

\small \begin{gather}
SFM_{(N,N,K)} = \begin{bmatrix}
\begin{bmatrix}
b_{(1,1,1)} & \cdots & b_{(1,N,1)} \\
b_{(2,1,1)} & \cdots & b_{(2,N,1)} \\
\vdots & \ddots & \vdots \\
b_{(N,1,1)} & \cdots & b_{(N,N,1)}
\end{bmatrix} \\
\vdots \\
\begin{bmatrix}
b_{(1,1,K)} & \cdots & b_{(1,N,K)} \\
b_{(2,1,K)} & \cdots & b_{(2,N,K)} \\
\vdots & \ddots & \vdots \\
b_{(N,1,K)} & \cdots & b_{(N,N,K)}
\end{bmatrix} 
\end{bmatrix} 
\in \mathbb{Z}^{N \times N \times K}_{0+},
\end{gather} \normalsize

being $b_{(i,j,k)}$ a category-pair inter-object relationship bin, $N$ the number of object categories, and $K$ the number of distance bins that represents the distances in terms of how close or apart objects are between two pairs of object categories. Each bin ($b_{(i,j,k)}$) is calculated as follows:

\small \begin{equation}
\label{eq:bin}
b_{(i,j,k)}=\sum_{m=1}^{|C^{[i]}|}{\sum_{n=1}^{|C^{[j]}|}{  }} f(C^{[i]}_n,C^{[j]}_m,k)
\end{equation} \normalsize
with
\small \begin{equation}
f(C_A,C_B,k))=\left\{\begin{matrix}
1 & \textup{if }\rho\frac{|C_A-C_B|}{d_{max}} =k , k \in [1,K]\\
0 & \textup{otherwise }
\end{matrix}\right.
\end{equation} \normalsize
where $C^{[i]}$ and $C^{[j]}$ represent the recognized objects' bounding box coordinates for each object category ($i$ and $j$), $d_{max}$ is the image's Euclidean distance (image's diagonal distance), and $\rho$ is a scaling factor for the number of distance bins. A zero inter-object relationship bin is assigned (i.e. $b(...) = 0$ (\ref{eq:bin}) for no detected object-categories.

%\subsubsection*{SFM-CNN Architecture}
To further exploit and extract non-linear patterns from the SFM, a CNN-based architecture \cite{gsf2appV2_IROS21}, SFM-CNN, as shown in Fig. \ref{fig:gos2f2app_overview}, is used. It is composed of three 2D convolutional layers (with 64, 128, and 64 output channels, respectively) followed by an FC layer. Each convolutional layer is composed of a kernel size of 3, a stride value of 1, and a padding value of 1. Furthermore, between the first two 2D convolutional layers, a max pooling layer to decrease the feature space size by half is applied. All layers use the ReLU as the activation function.

\subsection{Global Branch}
Although semantic-based features are important to represent a scene they are not enough, since the previous features only characterize a set of relevant segmentation-categories/objects and do not take into account global aspects of the scene such as the background or the building structure/characteristics. 
Hence, to achieve a more complete feature representation of a scene, global features are extracted and learned from RGB images through a state-of-the-art deep-learning-based feature extraction approach (backbone), as represented in the top branch of the proposed GOS$^2$F$^2$App (see Fig. \ref{fig:gos2f2app_overview}). Similar to \cite{gsf2appV2_IROS21, gs2f2app_ICRA23}, to show that the proposed semantic features, segmentation-based and object-based features, do not have a dependency pattern with a specific backbone, different backbone networks are evaluated in this work. Also, the use of multiple backbones allows to exploit the effectiveness and the capability that each backbone may have on extracting rich and complex features for indoor scene classification tasks. Thus, the following five state-of-the-art CNN-based backbones were evaluated in this work: ResNet50 \cite{resnet_2016}, ResNet101 \cite{resnet_2016}, DenseNet \cite{densenet_2017}, ResNeXt50 (32$\times$4d) \cite{resNext_2017}, and ConvNeXt-B \cite{convNext_2022}.
Moreover, the recently proposed transformer-based Swin-B \cite{swin_2021} network is also used as a backbone network.
The global branch is also composed of an FC layer that receives the backbone's output feature space as input. The FC layer uses the ReLU as the activation function.

\subsection{Feature Fusion}
\label{subsec:feature_fusion}

\begin{figure}[!tb] 
    \centering 
    \includegraphics[width=0.95\columnwidth] {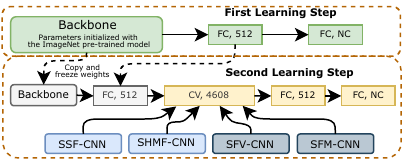} 
    \caption{Two-step learning of the GOS$^2$F$^2$App. In the first step, the global branch parameters, composed of the backbone network and an FC layer, are trained. The backbone network parameters are initialized using the ImageNet pre-trained model. In the second learning step, the global branch uses the parameters trained in the first step, and only the remaining parameters (segmentation branch, object-based branch and feature fusion parameters) are trained.}
    \label{fig:learning}
\end{figure}

To exploit the correlation between the different types of features used in the proposed approach, a two-step learning technique, as presented in Fig. \ref{fig:learning}, is used. In the first learning step, only the global branch parameters are trained, originating a global feature-based model. The parameters are initialized by copying the ImageNet pre-trained model' parameters. In the second learning step, the global branch parameters, trained in the first step, are frozen, and only the semantic-based CNN's parameters (segmentation and object-based branches' parameters) are trained, as well as the feature concatenation and scene prediction parameters. To obtain a more complete feature representation, the outputs' feature vector of each type of features are concatenated.

%%%%%%%%%%%%%%%%%%%%%%%%%%%%%%%%%%%%%%%%%%%%%%%%%%%%%%%%%%%%%%%%%%%%%%%%%%%%%%%%
\section{Experiments}
\label{sec:experiments}

The proposed approach was evaluated on two indoor scene classification benchmark datasets: SUN RGB-D \cite{sun_dataset} and NYU Depth Dataset v2 \cite{nyu_dataset}. Furthermore, a comparison between the achieved and reported state-of-the-art results (as reported in the original papers) was performed. For a more thorough evaluation of the proposed approach, a detailed ablation study, with extensive experiments on both datasets, is also presented.

\subsection{Datasets}
%The \textbf{SUN RGB-D dataset} has available 10355 RGB-D image pairs for 40 different indoor scene categories. The dataset was captured using four different RGB-D sensors: Kinect v1, Kinect v2, RealSense, and Asus Xtion. Following the public split \cite{sun_dataset}, scene categories with less than 80 images were discarded, resulting in 9504 RGB-D image pairs with 19 different indoor scene categories. Furthermore, the dataset was split in 4845 training images and 4659 testing images. Also, the dataset has available two semantic segmentation masks for each scene image, one containing all segmentation-categories (6585) and the other containing 37 segmentation-categories, being the last one used in this work.

The \textbf{SUN RGB-D dataset} \cite{sun_dataset} has available 9504 RGB-D image pairs with 19 different indoor scene categories with a public training/testing split of 4845/4659 images, respectively. Also, the dataset has available semantic segmentation masks, for a total of 37 segmentation-categories.

%The \textbf{NYU Depth dataset v2 (NYUv2)} contains 1449 RGB-D image pairs organized into 27 different indoor scene categories. However, following the benchmark split \cite{nyu_data_split}, the categories were reorganized into 10 scene categories, including 9 most common categories and the "other" category that contains the remaining categories. Furthermore, the split has 795 training images and 654 testing images. Also, the dataset provides two semantic segmentation masks for each scene image, one using all segmentation-categories (894) and the other using 40 segmentation-categories, being the last one used in this work.

The \textbf{NYU Depth dataset v2 (NYUv2)} \cite{nyu_dataset} contains 1449 RGB-D image pairs organized into 10 different indoor scene categories with a public training/testing split \cite{nyu_data_split} of 795/654 images, respectively. Also, the dataset has available semantic segmentation masks, for a total of 40 segmentation-categories.

\subsection{Implementation Details}
The proposed approach was implemented and evaluated using the Python 3.10.4 programming language, the PyTorch framework (version 1.12.1), and performed using an AMD Ryzen 9-5900X-@-3.7 GHz, 64GB RAM, and an Nvidia RTX 3090 GPU.

To train the segmentation network, DeepLabv3+ \cite{DeepLabv3_2018}, its encoder's parameters were initialized with the ImageNet pre-trained model's parameters and fine-tuned over 100 epochs using the AdamW optimizer with a learning rate of 10$^{-3}$ and a weight decay of 0.05. Furthermore, a cosine annealing schedule to make partial warm restarts of the learning rate was used. The network receives, as input, a 224$\times$224 RGB image and outputs a 224$\times$224 segmentation mask.
The generalization capabilities of the segmentation network are not included in the scope of this work, being the main focus on assessing the performance of the proposed semantic feature representation in the GOS$^2$F$^2$App. Hence, according to \cite{Seg_benchmark}, the state-of-the-art segmentation result obtained on the SUN RGB-D and NYUv2 datasets is $\approx 50 \%$. Therefore, to obtain segmentation data with higher performance to evaluate the proposed approach, both training and testing sets of the datasets were used in the training phase.

To detect and classify the objects available in the scene, the YOLOv3 network using the COCO \cite{coco_data} pre-trained model (able to detect eighty different object categories) with a confidence threshold of 0.2 \cite{gsf2app_ICARSC20} was used. The YOLOv3 receives, as input, a 416$\times$416 RGB image and outputs detected object bounding boxes coordinates and object category predictions.

The proposed approach, GOS$^2$F$^2$App, was trained in two steps, as described in \ref{subsec:feature_fusion}. In both training steps, GOS$^2$F$^2$App's parameters were fine-tuned over 100 epochs using the Adam optimizer with a learning rate of 10$^{-4}$, a weight decay rate of 5$\times$10$^{-4}$, and a batch size of 32. The GOS$^2$F$^2$App backbone's network receives, as input, a 224$\times$224 RGB image. For the SFM features, based on results reported in \cite{gsf2appV2_IROS21}, a number of distance bins of 3 ($K=3$) and a scaling factor of 3 ($\rho=3$) were used.

\subsection{Comparison with SOTA Methods}

To show the effectiveness of the proposed approach, the results achieved with the proposed method are compared with previous state-of-the-art results, as shown in Table \ref{tab:state_of_the_art_results}. The proposed GOS$^2$F$^2$App obtained 63.7\% and 80.1\% accuracy values on the SUN RGB-D and NYUv2 datasets, respectively, which are, to the best of our knowledge, the highest reported state-of-the-art results. Overall, the proposed approach is an enhanced version of \cite{gsf2appV2_IROS21,gs2f2app_ICRA23,gsf2app_ICARSC20}. In \cite{gsf2app_ICARSC20}, the occurrence of the objects recognized over the scenes is exploited and concatenated with VGG16-based global features. Then, to enhance the object-based distribution of the scenes, in addition to the objects' occurrence, a 2D-based spatial objects' distribution using inter-object distance relationships was exploited \cite{gsf2appV2_IROS21}. In \cite{gs2f2app_ICRA23}, a significant performance improvement was reached by exploiting a 2D-based spatial segmentation-categories' distribution using semantic segmentation masks as a source of semantic information. The approach proposed in this paper integrates the aforementioned approaches with a Hu-moments-based objects' shape characterization, which allows to obtain a more discriminative and complete feature representation of the scenes, leading to a significant performance improvement. On the other hand, significant positive performances were also obtained by \cite{Montoro_2021,caglayan_2022,song_oor_2020,du_pyramid_2021,Ayub_centroid_2020, Seichter_2022}.

\setlength{\tabcolsep}{4pt}
\begin{table}[!tb]
    \renewcommand{\arraystretch}{1.1}
    \centering
    \small
    \caption{Accuracy(\%) comparison with state-of-the-art methods.}
    \label{tab:state_of_the_art_results}
    \begin{tabular}{lcccc}
    \noalign{\hrule height 1pt} \hline
    \multirow{3}{*}{Method} & \multicolumn{4}{c}{Accuracy (\%)}                            \\ \cline{2-5} 
                            & \multicolumn{2}{c|}{SUN RGB-D}   & \multicolumn{2}{c}{NYUv2} \\ \cline{2-5} 
                            & RGB & \multicolumn{1}{c|}{RGB-D} & RGB        & RGB-D        \\ \hline
    Song \textit{et al.} \cite{song_2017} (2017) & 43.5   & \multicolumn{1}{c|}{54.0} & 57.3 & 66.9 \\
    Li \textit{et al.} \cite{df2net_2018} (2018) & 46.3   & \multicolumn{1}{c|}{54.8} & 61.1 & 65.4 \\
    Cai \& Shao \cite{Cai_2019} (2019)  & 46.4  & \multicolumn{1}{c|}{48.7} & - & -     \\
    Xiong \textit{et al.} \cite{xiong_2019} (2019) & -   & \multicolumn{1}{c|}{55.9} & 53.5  &  67.8 \\
    Li \textit{et al.} \cite{mapnet_2019} (2019)  & 46.0  & \multicolumn{1}{c|}{56.2} & -  & 67.7 \\
    Yuan \textit{et al.} \cite{Yuan_2019} (2019)  & 45.7  & \multicolumn{1}{c|}{55.1} & 55.4 & 67.2 \\
    Song \textit{et al.} \cite{Song_2019} (2019)  & -  & \multicolumn{1}{c|}{53.8}  & - &  67.5  \\
    Song \textit{et al.} \cite{song_oor_2020} (2020) & 50.5  & \multicolumn{1}{c|}{55.5} & 64.2  & 67.4 \\
    Pereira \textit{et al.} \cite{gsf2app_ICARSC20} (2020)  & 55.3 & \multicolumn{1}{c|}{-}  & 70.6  & -  \\
    Xiong \textit{et al.} \cite{xiong_MSN_2020} (2020) & -  & \multicolumn{1}{c|}{56.2} & 53.5 & 68.1 \\
    Ayub \& Wagner \cite{Ayub_centroid_2020} (2020) & 48.8  & \multicolumn{1}{c|}{59.5} & 66.4 & 70.9 \\
    Xiong \textit{et al.} \cite{Xiong_2021} (2021) & - & \multicolumn{1}{c|}{57.3} & 61.2 & 69.3 \\
    Pereira \textit{et al.} \cite{gsf2appV2_IROS21} (2021) & 58.1 & \multicolumn{1}{c|}{-} & 73.1 & - \\
    Montoro \& Hidalgo \cite{Montoro_2021} (2021) & 56.4  & \multicolumn{1}{c|}{58.6} & 67.8  & 75.1 \\
    Du \textit{et al.} \cite{du_pyramid_2021} (2021) & 53.8 & \multicolumn{1}{c|}{58.5} & 66.1 & 71.8  \\
    Caglayan \textit{et al.} \cite{caglayan_2022} (2022) & 58.5 & \multicolumn{1}{c|}{60.7} &  -  & - \\
    Seichter \textit{et al.} \cite{Seichter_2022} (2022) & - & \multicolumn{1}{c|}{61.8} &  -  & 76.5 \\
    Zhou \textit{et al.} \cite{zhou_2023} (2023) & 59.5 & \multicolumn{1}{c|}{-} & - & -   \\
    Pereira \textit{et al.} \cite{gs2f2app_ICRA23} (2024) & 62.3 & \multicolumn{1}{c|}{-} & 77.8 & -  \\ \hline
    GOS$^2$F$^2$App (ours)    & \textbf{63.7}  & \multicolumn{1}{c|}{-}  & \textbf{80.1} & -   \\
    \noalign{\hrule height 1pt} \hline
\end{tabular}
\end{table}

Caglayan \textit{et al.} \cite{caglayan_2022} extracts CNN-based RGB-D features at multiple levels, which were mapped into high-level representations using a randomized structure of recursive neural networks. Such approach allows the generation of highly discriminative deep-learning-based features. However, it does not take into consideration any semantic information about the scenes. Montoro \& Hidalgo \cite{Montoro_2021} combined 2D CNN-based texture features with 3D geometric features that, indirectly, end up exploiting object-based features.
Ayub \& Wagner \cite{Ayub_centroid_2020} developed a centroid-based concept learning that generates clusters and centroid pairs for different scene categories. They used CNN-based RGB-D feature maps to compare distances to centroid pairs in the scene category and update them accordingly. This approach can obtain a promising discriminative feature representation of the scene but may face memory and computation issues with a large number of images and categories.

The proposed work shares common ideas with other approaches such as exploiting semantic information to improve the indoor scene classification performance.
Seichter \cite{Seichter_2022} \textit{et al.} proposed an RGB-D multi-task network aimed at scene classification, instance orientation, and semantic and instance segmentation. The network uses a shared backbone network with task-specific heads, enabling the backbone network to learn how to extract global and semantic features.
Song \textit{et al.} \cite{song_2017} developed an object-to-object relationship representation based on the detected objects in the scene to generate spatial layouts that can characterize indoor scenes. 
Song \textit{et al.} \cite{song_oor_2020} also exploited two different object-based representations: co-occurrence frequency of object-to-object relationships and sequential occurrence of object-to-object relationships. The first representation focuses on exploiting the object-based scene's layout while the second representation focuses on describing the scene in a sequential form. Yuan \textit{et al.} \cite{Yuan_2019} focuses on detecting key objects or object parts from RGB-D feature maps and then exploiting their relationships through CNN-based graph networks, which can find patterns in non-structured data. 
Although the use of object detector networks can enhance indoor scene classification performance, they may lack object bounding box annotations for the indoor scene classification task, requiring a pre-trained model in other datasets that can lead to incorrect predictions.
Additionally, these approaches do not exploit objects' shapes, which can provide valuable semantic information that can disentangle feature representations of indoor scenes.
As a way to complement the object-based information provided by object detector networks, the proposed approach uses semantic segmentation masks to obtain a Hu-moments-based segmentation-categories' shape characterization and a 2D-based spatial segmentation-categories' distribution that led to a significant performance improvement, as shown in Table \ref{tab:state_of_the_art_results}.

\subsection{Ablation Study}
%\subsubsection{Ablation Study}
To better gauge the influence of the different features exploited in the proposed approach, the following intermediate evaluations were performed: semantic features (segmentation-based and object-based features, e.g. SHMFs, SSFs+SHMFs (Sb)); global features-only (RGB); global features combined with SHMFs (RGB+SHMFs); global features combined with SHMFs and SSFs (RGB +Sb); global features combined with SHMFs and both object-based features (RGB+SHMFs+Ob); and global features extracted by different backbone networks (e.g. ResNet50 \cite{resnet_2016}, Swin-B \cite{swin_2021}). Moreover, a comparison between the proposed approach and previous works, GSF$^2$AppV2 \cite{gsf2appV2_IROS21} and GS$^2$F$^2$App \cite{gs2f2app_ICRA23}, was also carried out.

To assess the influence that segmentation models (SMs) with different accuracy rates may have on the proposed approach, three different SMs (S20, S50, and S75) with different segmentation overall accuracies, were also used in the intermediate evaluations. S20 represents a SM with approximately 20\% of overall accuracy, S50 a SM with an overall accuracy of around 50\%, and S75 a SM with an overall accuracy of 75\%. Based on the state-of-the-art segmentation results \cite{Seg_benchmark}, S50 was considered for the benchmark result. %, as used in \cite{gs2f2app_ICRA23}.

%\subsubsection{SUN RGB-D Dataset}
\subsubsection{Semantic Features} % Semantic Features-only Evaluation

%Table \ref{tab:SFonly_results} presents the accuracy values, for the semantic features-only evaluation, obtained on the SUN RGB-D dataset. Reported results show that the best performance was achieved by using the four semantic features (SHMFs, SSFs, SFV, and SFM), which means that combining the four semantic features generates a more descriptive feature representation of the scene when compared to the use of other semantic feature combinations. 
%Moreover, a significant performance improvement was obtained by using the four semantic features instead of using segmentation-only features (SSF-CNN+SHMF-CNN) or object-based features (SFV-CNN+SFM-CNN), which shows that the two types of semantic features can complement each other in order to achieve a better semantic feature representation of the indoor scene. 
%Comparing the reported results using the segmentation-based features, the proposed SHMFs (SHMF-CNN) achieved lower accuracy values when compared to the SSFs (SSF-CNN). This may happen since SHMFs are more focused on the segmentation-categories' shape, while SSFs are more focused on the 2D spatial distribution of the segmentation-categories. However, a significant improvement is achieved when SHMFs are combined with object-based features (SHMF-CNN+SFV-CNN+SFM-CNN).

Table \ref{tab:SFonly_results} presents the accuracy values, for the evaluation of the semantic features, obtained on the SUN RGB-D and NYUv2 datasets. Combining the four semantic features (Sb+Ob) achieved the highest accuracy values, 48.8\% on SUN RGB-D dataset and 70.0\% on NYUv2 dataset, which means that a more descriptive feature representation of the scene is obtained when compared to the use of other semantic feature combinations. Such promising results are getting closer to those achieved by the backbone networks, specially on the NYUv2 dataset (RGB in Table \ref{tab:results}), which highlights the capability of the semantic features in generating a descriptive feature representation that leads to successful scene predictions.
Moreover, a significant performance improvement was obtained by Sb+Ob features instead of using Sb features or Ob features, showing that the two types of semantic features can complement each other in order to achieve a better semantic feature representation of the indoor scene.
Comparing the reported results using the segmentation-based features, the proposed SHMFs achieved lower accuracy values when compared to the SSFs. This may happen since SHMFs are more focused on the segmentation-categories' shape, while SSFs are more focused on the 2D spatial distribution of the segmentation-categories, being the latest able to achieve a better feature representation of the indoor scenes. However, a significant improvement is achieved when SHMFs are combined with object-based features (SHMFs+Ob).

\setlength{\tabcolsep}{2.7pt}
\begin{table}[!tb]
    \renewcommand{\arraystretch}{1.1}
    \centering
    \small
    \caption{Accuracy (\%) values for the evaluation of the semantic features.}
    \label{tab:SFonly_results}
    \begin{tabular}{lcccccccc}
     \noalign{\hrule height 1pt} \hline
    \multirow{4}{*}{Features}                      & \multicolumn{8}{c}{Accuracy (\%)}                                                                                     \\ \cline{2-9} 
                                                   & \multicolumn{4}{c|}{SUN RGB-D}                                    & \multicolumn{4}{c}{NYUv2}                         \\ \cline{2-9} 
                                                   & \multicolumn{3}{c}{w\textbackslash SM} & \multicolumn{1}{c|}{\multirow{2}{*}{w\textbackslash o SM}} & \multicolumn{3}{c}{w\textbackslash SM}      & \multirow{2}{*}{w\textbackslash o SM} \\ \cline{2-4} \cline{6-8}
                                                   & S20    & S50   & S75   & \multicolumn{1}{c|}{}                    & S20  & S50  & S75           &                     \\ \hline
    SSFs \cite{gs2f2app_ICRA23}  & 34.6   & 43.5  & 44.5  & \multicolumn{1}{c|}{-}                   & 56.4 & 67.3 & 70.0          & -                   \\
    SHMFs                                          & 29.9   & 41.5  & 43.3  & \multicolumn{1}{c|}{-}                   & 53.6 & 65.9 & 65.9          & -                   \\
    Sb                                             & 33.3   & 43.6  & 44.7  & \multicolumn{1}{c|}{-}                   & 54.8 & 68.9 & \textbf{70.4} & -                   \\
    SFV                                            & -      & -     & -     & \multicolumn{1}{c|}{40.5}                & -    & -    & -             & 62.8                \\
    SFM   & -      & -     & -     & \multicolumn{1}{c|}{39.2}                & -    & -    & -             & 58.5                \\
    Ob                                             & -      & -     & -     & \multicolumn{1}{c|}{\textbf{41.7}}       & -    & -    & -             & \textbf{64.1}       \\
    SHMFs+Ob                                       & 41.5   & 48.3  & 49.0  & \multicolumn{1}{c|}{-}                   & 58.4 & 68.0 & 68.3          & -                   \\
    Sb+Ob & \textbf{41.9} & \textbf{48.8} & \textbf{49.4} & \multicolumn{1}{c|}{-} & \textbf{59.6} & \textbf{70.0} & \textbf{70.4} & - \\  \noalign{\hrule height 1pt} \hline
    \end{tabular}
\end{table}

Similar to the reported in \cite{gs2f2app_ICRA23}, attained results show that using segmentation masks generated by the S50 model led to a significant improvement over the use of segmentation masks generated by the S20 model, which are again improved by using segmentation masks generated by the S75 model. Therefore, as expected, as better segmentation models were used, higher overall performances were reached, which shows that the performance of the segmentation model may have considerable relevance in the overall scene classification performance.
Using the S50 model as a benchmark result, the segmentation-based features achieved higher accuracy values than object-based features. Such result was expected, since segmentation masks provide information about the whole scene, while object bounding boxes only provide information about the detected objects.

\subsubsection{Global and Semantic Feature Fusion}

%Tables \ref{tab:sun_results} and \ref{tab:nyu_results}
Table \ref{tab:results} shows the accuracy values obtained by the proposed approach and intermediate evaluations of combining global features with semantic features using different global feature extraction approaches as backbone/baseline on the SUN RGB-D and NYUv2 datasets, respectively. 
Highlight values are presented in form of bold and underline values.
Bold values highlight, per feature combination, the backbone that achieved the highest result, while underline values highlight, per each backbone, the feature combination that reached the highest value using segmentation features extracted from the S50 model.

%Table \ref{tab:sun_results} shows the accuracy values achieved on the SUN RGB-D dataset for the proposed approach and intermediate evaluation of combining global features with semantic features using different global feature extraction approaches as backbone/baseline. 
Overall, promising results were attained by the GOS$^2$F$^2$App, which presents a significant improvement when compared with the baseline results (RGB). 
%Also, the proposed approach presents a significant performance improvement over the previous works \cite{gsf2appV2_IROS21,gs2f2app_ICRA23} 
Also, using the same backbone network as in \cite{gsf2appV2_IROS21,gs2f2app_ICRA23} and using semantic features obtained from segmentation masks generated from the S50 model, the GOS$^2$F$^2$App presents a significant performance improvement over the previous works \cite{gsf2appV2_IROS21,gs2f2app_ICRA23}, showing the positive impact of the proposed approach.
Reported results attained by the GOS$^2$F$^2$App also show a dependency pattern regarding the quality of the semantic segmentation masks. Using segmentation-based features generated from segmentation masks obtained from the S20 model achieved lower accuracy values, while using segmentation-based features extracted from segmentation masks obtained from the S75 model attained higher accuracy values. Such results were expected, since segmentation masks generated by the S75 model present a better representation of the scene than segmentation masks generated by the S50 or the S20 models. Hence, features generated from segmentation masks provided by the S75 model allow to obtain a more meaningful and descriptive feature representation of the scene, which leads to better overall performances.

\setlength{\tabcolsep}{2.2pt}
\begin{table*}[!tb]
    \renewcommand{\arraystretch}{1.1}
    \centering
    \small
    \caption{Global and semantic feature fusion results obtained on both SUN RGB-D and NYUv2 datasets.}
    \label{tab:results}
    \begin{tabular}{lcccccccccccccccccc}
        \noalign{\hrule height 1pt} \hline
        \multirow{4}{*}{Backbone} & \multicolumn{18}{c}{Accuracy (\%)} \\ \cline{2-19} 
         & \multicolumn{9}{c|}{SUN RGB-D} & \multicolumn{9}{c}{NYUv2} \\ \cline{2-19} 
         & \multirow{2}{*}{RGB} & \begin{tabular}[c]{@{}c@{}}RGB+\\ SSFs\\ \cite{gs2f2app_ICRA23}\end{tabular} & \begin{tabular}[c]{@{}c@{}}RGB+\\ SHMFs\end{tabular} & \begin{tabular}[c]{@{}c@{}}RGB\\ +Sb\end{tabular} & \begin{tabular}[c]{@{}c@{}}RGB\\ +Ob\\ \cite{gsf2appV2_IROS21}\end{tabular} & \begin{tabular}[c]{@{}c@{}}RGB+\\ SHMFs\\ +Ob\end{tabular} & \multicolumn{3}{c|}{GOS$^2$F$^2$App} & \multirow{2}{*}{RGB} & \begin{tabular}[c]{@{}c@{}}RGB+\\ SSFs\\ \cite{gs2f2app_ICRA23}\end{tabular} & \begin{tabular}[c]{@{}c@{}}RGB+\\ SHMFs\end{tabular} & \begin{tabular}[c]{@{}c@{}}RGB\\ +Sb\end{tabular} & \begin{tabular}[c]{@{}c@{}}RGB\\ +Ob\\ \cite{gsf2appV2_IROS21}\end{tabular} & \begin{tabular}[c]{@{}c@{}}RGB+\\ SHMFs\\ +Ob\end{tabular} & \multicolumn{3}{c}{GOS$^2$F$^2$App} \\ \cline{3-10} \cline{12-19} 
         &  & S50 & S50 & S50 & - & S50 & S20 & S50 & \multicolumn{1}{c|}{S75} &  & S50 & S50 & S50 &  & S50 & S20 & S50 & S75 \\ \hline
        ResNet50 \cite{resnet_2016} & 56.8 & 59.8 & 60.4 & 60.7 & 60.3 & 61.0 & 60.6 & \underline{61.5} & \multicolumn{1}{c|}{61.5} & 70.8 & 75.5 & 76.3 & 76.7 & 75.2 & 77.0 & 74.2 & \underline{77.5} & 78.1 \\
        ResNet101 \cite{resnet_2016} & 56.3 & 59.2 & 59.6 & 59.8 & 59.3 & 60.3 & 59.6 & \underline{60.6} & \multicolumn{1}{c|}{60.7} & 70.7 & 74.9 & 76.6 & 77.0 & 74.7 & 77.1 & 73.5 & \underline{77.5} & 78.3 \\
        DenseNet \cite{densenet_2017} & 57.1 & 58.9 & 59.8 & 60.3 & 59.7 & 60.8 & 59.4 & \underline{61.1} & \multicolumn{1}{c|}{61.6} & 70.8 & 75.8 & 76.3 & 76.9 & 73.4 & 77.1 & 72.6 & \underline{77.8} & 77.6 \\
        ResNeXt50 \cite{resNext_2017} & 56.9 & 62.1 & 62.5 & 63.0 & \textbf{62.5} & 63.3 & \textbf{63.0} & \underline{63.6} & \multicolumn{1}{c|}{64.0} & 71.7 & \textbf{77.8} & \textbf{78.4} & \textbf{79.0} & \textbf{76.6} & \textbf{79.2} & \textbf{77.2} & \underline{\textbf{80.1}} & \textbf{81.2} \\
        Swin-B \cite{swin_2021} & 57.3 & 61.1 & 61.8 & 62.1 & 61.3 & 62.6 & 61.8 & \underline{63.0} & \multicolumn{1}{c|}{62.9} & 71.5 & 76.1 & 76.7 & 77.6 & 73.4 & 78.0 & 74.2 & \underline{79.0} & 79.2 \\
        ConvNeXt-B \cite{convNext_2022} & \textbf{58.3} & \textbf{62.3} & \textbf{62.8} & \textbf{63.2} & \textbf{62.5} & \textbf{63.5} & 62.6 & \underline{\textbf{63.7}} & \multicolumn{1}{c|}{\textbf{64.1}} & \textbf{72.7} & 76.7 & 77.1 & 77.7 & 75.0 & 78.0 & 74.7 & \underline{78.6} & 78.9 \\ \noalign{\hrule height 1pt} \hline
    \end{tabular}
\end{table*}

Regarding the performed intermediate evaluations, promising results were also achieved. Combining global features with the proposed SHMFs (RGB+ SHMFs) outperformed the baseline results. Moreover, regardless the backbone in use, combining global features with SHMFs also outperformed the combination of global features with SSFs (RGB+ SSFs), which shows the effectiveness of the proposed semantic features to obtain a more descriptive feature representation of the scene over the previous work \cite{gs2f2app_ICRA23}. 
Furthermore, reported results show that the accuracy values are increasing as more types of semantic features are combined with global features, i.e., combining global features with the segmentation-based features (RGB+Sb) attained a higher accuracy value than using global features with SHMFs, which was again improved by combining global features with the SHMFs and object-based features (RGB+SHMFs+Ob). The aforementioned behavior of the accuracy values was expected, since aggregating more types of semantic features allows to obtain a more descriptive feature representation of the scene that, when also combined with global features, leads to better performance.

Moreover, promising results regarding the backbone performances (RGB) are also reported. On the SUN RGB-D dataset, the ResNet101 achieved the lowest accuracy value, 56.3\%, while the ConvNeXt-B attained the highest accuracy value, 58.3\%. The remaining backbones reached similar performances, being the accuracy values ranging between 56.8\% and 57.3\%.
On the other hand, using segmentation-based features extracted from segmentation masks generated by the S50 model, the proposed approach using the ResNet101 and ConvNeXt-B networks as backbones reached 60.6\% and 63.7\% accuracy values, respectively. The accuracy values attained by the proposed approach using the remaining networks as backbone ranged between 61.1\% to 63.6\%. Therefore, comparing the obtained backbone results with the results achieved by the GOS$^2$F$^2$App, it is possible to conclude that the proposed approach does not have a backbone dependency based on the obtained backbone accuracy value. As expected, the overall performance of the GOS$^2$F$^2$App relies on the capability of the backbone to generate, a more descriptive and complete as possible, global feature space representation.

\begin{figure}[!tb]  
    \centering % trim={<left> <lower> <right> <upper>}
    \includegraphics[ trim={16.1cm 1.65cm 8.8cm 2.8cm}, clip, width=\linewidth] {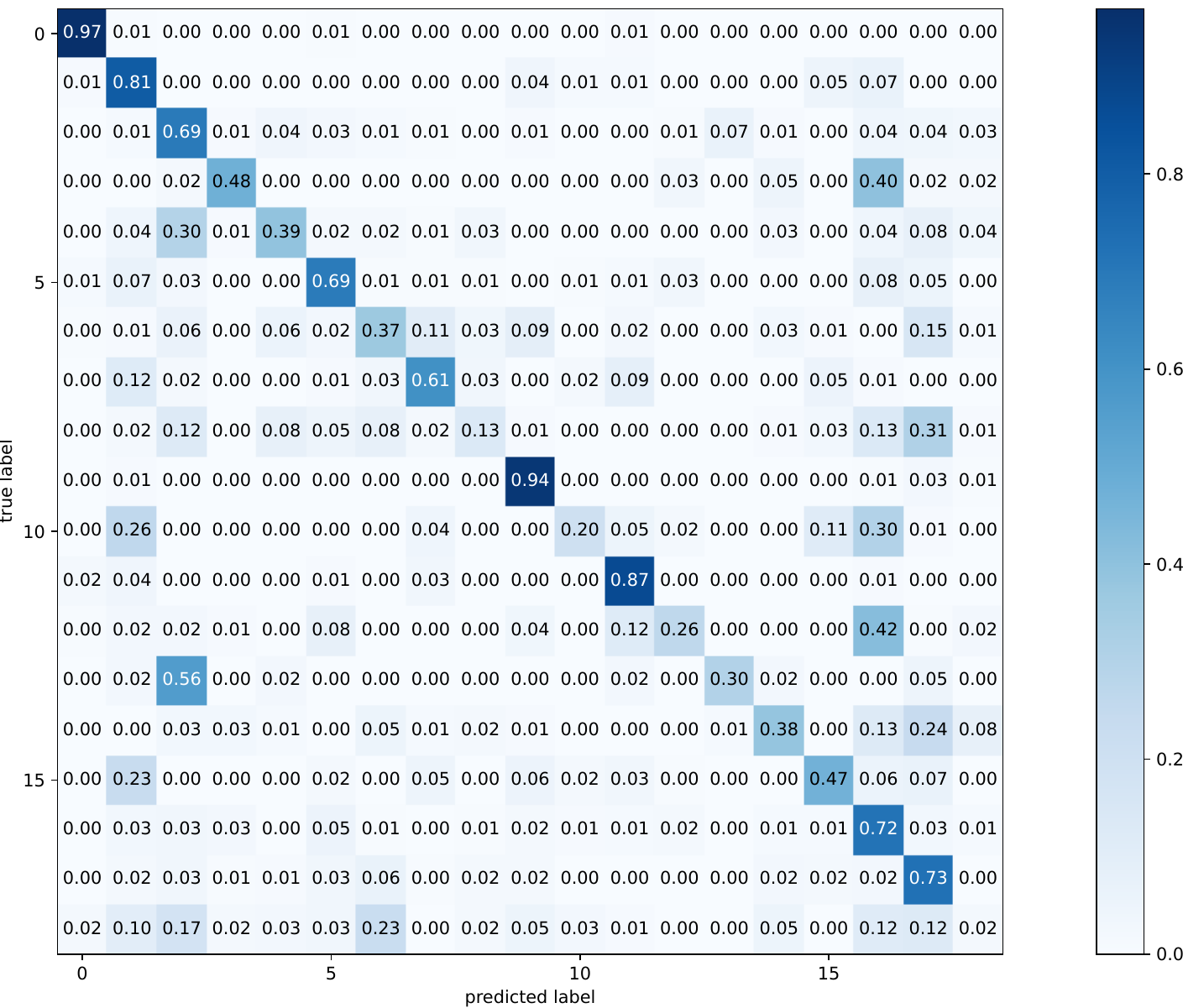}
    \caption{Confusion matrix of GOS$^2$F$^2$App on the SUN RGB-D dataset (Classes: 0 = bathroom, 1 = bedroom, 2 = classroom, 3 = computer room, 4 = conference room, 5 = corridor, 6 = dining area, 7 = dining room, 8 = discussion area, 9 = furniture store, 10 = home office, 11 = kitchen, 12 = lab, 13 = lecture theater , 14 = library, 15 = living room, 16 = office, 17 = rest space, 18 = study space).}
    \label{fig:sun_conf_matrix}
\end{figure}

\begin{figure*}[!tb]
	\centering
	\includegraphics[width=\linewidth]{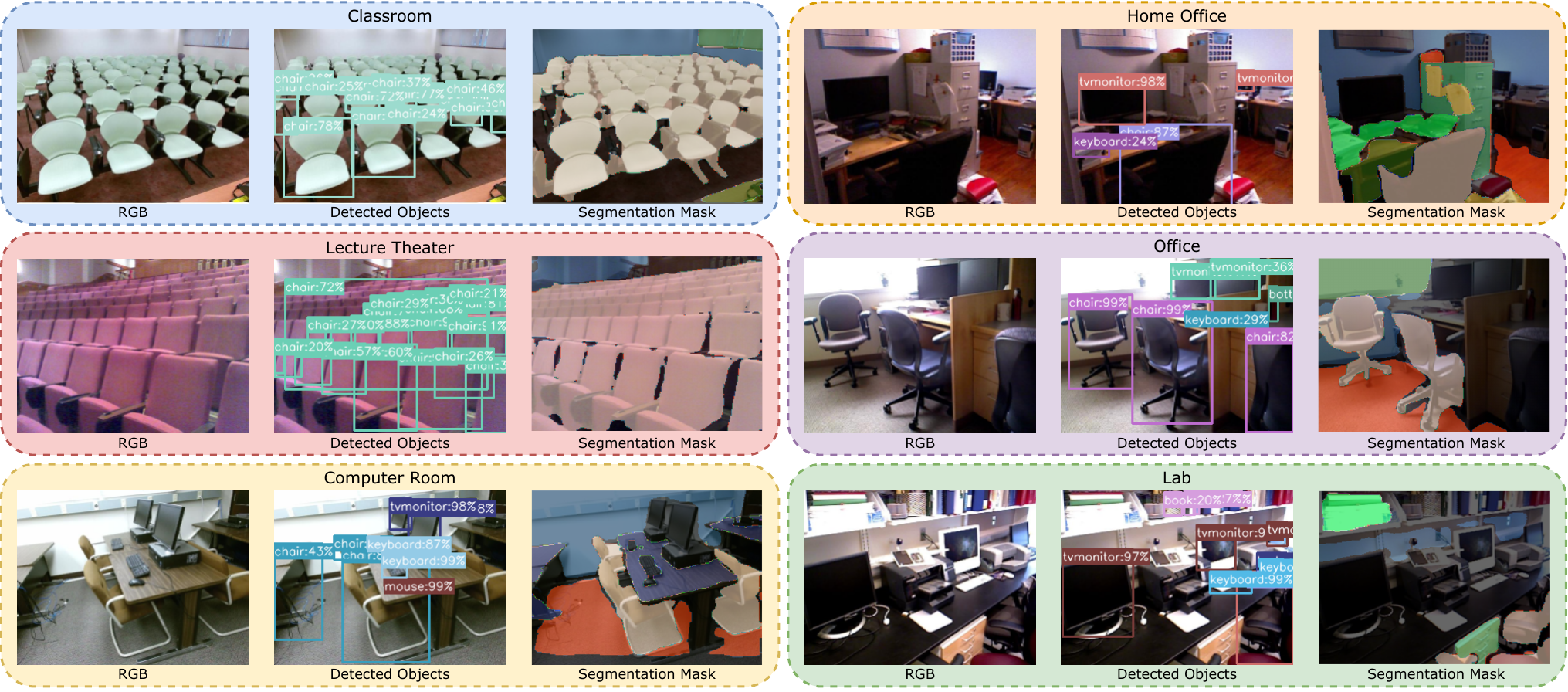}
	\caption{Misclassification examples of the GOS$^2$F$^2$App on the SUN RGB-D dataset. For each scene category, an RGB image is presented. Also, to present the high level of similarity between scene categories, an RGB image with object bounding boxes overlaid on detected objects and an RGB image with a segmentation mask generated by the S75 model superimposed, are presented.}
	\label{fig:misclassification}	
\end{figure*}

Using the S50 model to generate the segmentation-based features and the ConvNeXt-B as the backbone network, the proposed approach achieved the highest accuracy value on the SUN RGB-D dataset, 63.7\%, whose confusion matrix is shown in Fig. \ref{fig:sun_conf_matrix}. 
%It can be seen that the GOS$^2$F$^2$App was able to successfully predict a subset of categories: \textit{bathroom}, \textit{bedroom}, \textit{classroom}, \textit{corridor}, \textit{dining room}, \textit{furniture store}, \textit{kitchen}, \textit{office}, and \textit{rest space}. On the other hand, the proposed approach was not able to recognize the \textit{home office} and the \textit{lab}, which were classified as \textit{office}. Also, the \textit{home office} was misclassified as \textit{bedroom}. Furthermore, the \textit{discussion area} and the \textit{lecture theater} were misclassified as \textit{rest space} and \textit{classroom}, respectively. Despite the correct recognition of the \textit{computer room}, it is also misclassified as \textit{office}. Moreover, the GOS$^2$F$^2$App was not able to recognize the \textit{study space}, being misclassified as others categories, which shows that the proposed approach can not generate enough descriptive features about this scene category. 
It can be seen that the GOS$^2$F$^2$App was able to successfully predict a subset of categories. On the other hand, the GOS$^2$F$^2$App misclassified some scene categories.
Such misclassifications occur, as shown in Fig. \ref{fig:misclassification}, due to a high degree of inter-category similarity, where extracted global and semantic features, become very similar, which increases the difficult of reaching a successful scene category prediction.
However, comparing the reported confusion matrix with the confusion matrices reported in \cite{gsf2appV2_IROS21,gs2f2app_ICRA23}, a noticeable improvement achieved by the proposed approach is observed.

\begin{figure}[!tb] 
    \centering % trim={<left> <lower> <right> <upper>}
    \includegraphics[trim={1.8cm 0.3cm 2.2cm 1.35cm}, clip, width=0.75\linewidth]  {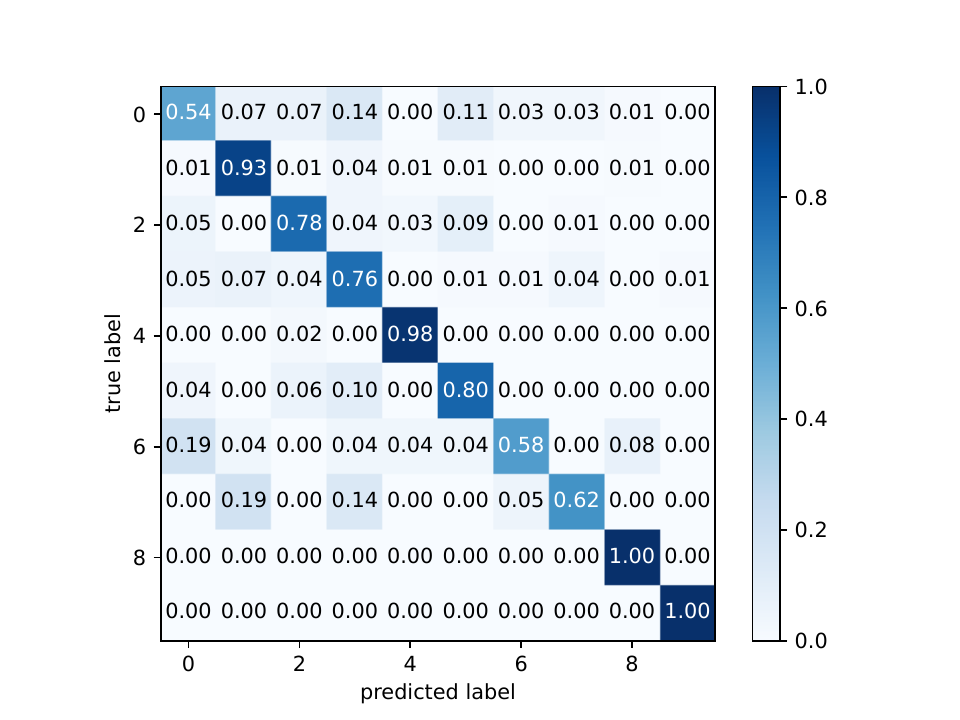} %0.8
    \caption{Confusion matrix of GOS$^2$F$^2$App on the NYUv2 dataset (Classes: 0 = others, 1 = bedroom, 2 = kitchen, 3= livingRoom, 4 = bathroom, 5 = diningRoom, 6 = office, 7 = homeOffice, 8 = classroom, 9= bookStorage).}
   \label{fig:nyu_conf_matrix}
\end{figure}

Comparing the results obtained for the backbone networks with those achieved by the GOS$^2$F$^2$App on the NYUv2 dataset, as shown in Table \ref{tab:results}, it is also possible to conclude that the proposed approach does not have a backbone dependency based on the obtained backbone accuracy value. Furthermore, 
%Based on the results obtained on the NYUv2 dataset, as shown in Table \ref{tab:nyu_results},
the ConvNeXt-B network reached the highest accuracy value on the backbone evaluation (RGB), 72.7\%, while on the remaining evaluations, the higher accuracy values were obtained using the ResNeXt50 as backbone. Hence, the GOS$^2$F$^2$App obtained the highest accuracy value using the ResNeXt50 network as the backbone, 80.1\%, whose confusion matrix is presented in Fig. \ref{fig:nyu_conf_matrix}. Overall, it can be seen that the majority of categories were well predicted with no major misclassifications. Comparing the reported confusion matrix with the confusion matrices reported in \cite{gsf2appV2_IROS21,gs2f2app_ICRA23}, a significant improvement can be observed.

\section{Conclusion}
\label{sec:conclusion}

In this paper, having in view a more complete and descriptive feature representation of indoor scenes, a novel Global, Object-based, and Semantic Segmentation Feature Fusion Approach (GOS$^2$F$^2$App), was proposed. Unlike other methods proposed in the literature, the GOS$^2$F$^2$App combines deep-learning-based global features with different types of semantic features such as object-based and segmentation-based features. While object-based features focus on how objects are distributed over the scenes, segmentation-based features focus on spatial distribution and shape-related features of the segmentation-categories. Hence, in this paper, a new approach that extracts Segmentation-based Hu-moments Features (SHMFs), which encodes a segmentation-categories' shape characterization was also proposed and integrated into the pipeline. The GOS$^2$F$^2$App was evaluated on two indoor scene benchmark datasets, SUN RGB-D and NYU Depth V2, achieving 63.7\% and 80.1\% accuracy values, respectively, which are, to the best of our knowledge, the best results reported in the literature. Such results highlight the effectiveness of the proposed approach in obtaining a more descriptive feature representation of the indoor scenes that led to scene classification improvements. Reported results showed that by combining both types of semantic features, a more meaningful and discriminative feature representation of the indoor scenes is obtained, leading to a significant performance improvement. On the other hand, reported results also showed that the overall performance of the GOS$^2$F$^2$App can be improved by improving the performance of the segmentation model from where segmentation-based features are extracted.
%On the other hand, reported results also showed that poor performance of the segmentation model used to extract the segmentation-based features negatively influences the overall performance of the GOS$^2$F$^2$App. 
In future work, attention layers to select key features from each branch can be exploited for a more discriminative feature fusion. Also, semantic features can be used as CNN-based graph network inputs to exploit further correlations between objects.

%%%%%%%%%%%%%%%%%%%%%%%%%%%%%%%%%%%%%%%%%%%%%%%%%%%%%%%%%%%%%%%%%%%%%%%%%%%%%%%%
\section*{Acknowledgments}
Ricardo Pereira and Tiago Barros have been supported by the Portuguese Foundation for Science and Technology (FCT) under the PhD grants with references SFRH/BD/148779/2019 and 2021.06492.BD, respectively. This work has been also supported by the FCT through grant UIDB/00048/2020 and partially funded by Agenda "GreenAuto: Green innovation for the Automotive Industry", with reference C644867037.

%%%%%%%%%%%%%%%%%%%%%%%%%%%%%%%%%%%%%%%%%%%%%%%%%%%%%%%%%%%%%%%%%%%%%%%%%%%%%%%%
\balance
\bibliographystyle{IEEEtran}
\bibliography{references}

\end{document}